\documentclass[11pt]{article}

\usepackage[preprint]{acl}

\usepackage{times}
\usepackage{latexsym}
\usepackage[T1]{fontenc}
\usepackage[utf8]{inputenc}
\usepackage{microtype}
\usepackage{inconsolata}

\usepackage{amsmath,amssymb}
\usepackage{graphicx}
\usepackage{subcaption}
\usepackage{booktabs}
\usepackage{multirow}
\usepackage{xcolor}

\usepackage{tikz}
\usetikzlibrary{positioning, arrows.meta, calc}

\newcommand{\Labs}{L22}      %
\newcommand{\Lform}{L36}     %
\newcommand{\Lbind}{L40}     %
\newcommand{\Lout}{L80}      %

\title{A Four-Stage Decomposition of Word-Problem Solving and \\ Mechanistic Fragility in LLM Math Reasoning}

\author{Zhongdi Qu\\
Cornell University \\
\texttt{zq84@cornell.edu} \\\And
Carla P. Gomes \\
Cornell University \\
\texttt{gomes@cs.cornell.edu}}

\begin{document}
\maketitle

\begin{abstract}
Large language models solve grade-school math word problems with high accuracy,
yet a single irrelevant clause inserted into the problem can collapse it. We
reconcile these observations with a mechanistic account. We show that the model's internal computation
decomposes into a four-stage sequential pipeline, Schema Abstraction, Operation
Planning, Operand Binding, and Computation, each stage producing a distinct
intermediate representation in an identifiable band of layers. Using the same scaffold to diagnose
distractor-induced failure, we localize the corruption to a single stage,
Operation Planning, implemented by a set of attention heads whose causal
role we validate bidirectionally. In short, we provide a mechanistic interpretation of math word problem reasoning in LLMs, and their failure when distracted.
\end{abstract}

\section{Introduction}
\label{sec:intro}

Two recent lines of work on LLM math word problem solving point in opposite directions. \citet{mirzadeh2025} show that inserting an irrelevant ``NoOp'' clause into a grade-school math word problem collapses accuracy, and conclude that LLM math reasoning is closer to surface pattern-matching than to genuine inference. \citet{cheng2025can} report a two-stage \emph{abstract-then-compute} pipeline in which the model first encodes the problem as an abstract schema and then computes the result numerically, evidence that LLMs do not rely on shallow pattern-matching shortcuts. We unify both findings into a single mechanistic account: LLMs do solve math word problems via an orderly pipeline, and NoOp fragility arises from the failure of a single stage in that pipeline, implemented by a specific set of attention heads. Pipeline structure and selective fragility coexist.

We present a four-stage pipeline: Schema Abstraction, Operation Planning, Operand Binding, and Computation. Each stage produces a distinct intermediate representation in the residual stream. We identify the stage boundaries with three independent methods: residual cosine similarity exposes them from layer-to-layer dynamics alone, per-category linear probes pinpoint what changes at each boundary, and cross-prompt activation patching tests which residual-stream changes causally determine the final answer. The three methods converge on the same four-stage decomposition.

We then use this scaffold to diagnose NoOp fragility. Activation patching localizes NoOp-induced failure to the Operation Planning stage. An engagement-anchored per-head Direct Logit Attribution identifies a compact set of attention heads that route through the distractor clause, splitting into subsets that write \emph{toward} and \emph{against} the wrong operation plan. Bidirectional attention scaling validates the heads causally: amplifying the anti-wrong-plan subset on NoOp prompts recovers a substantial fraction of the lost accuracy, while ablating either subset on data the model normally solves causes a catastrophic accuracy collapse. Together, these results identify the heads as the mechanistic instrument of Operation Planning, and NoOp fragility as what happens when this instrument is overwhelmed by distractor information.

\textbf{Our Contributions.}
(1) A mechanistically-grounded four-stage decomposition of LLM word-problem solving, and initial evidence that such decomposition generalizes to other reasoning domains. 
(2) A circuit-level diagnosis of NoOp fragility that localizes the failure to Operation Planning, implemented by a compact set of attention heads. 
(3) A training-free lever on the failure, i.e., a single global scalar that recovers $0.51$ of NoOp failures, locating where any future repair must act.\footnote{Accepted to Findings of EMNLP 2026. Code and datasets are available at \url{https://github.com/deliaqu/llm-reasoning-decomposed}.}

\section{Background and Related Work}
\label{sec:background}
\subsection{Fragility of LLM math reasoning}\label{sec:background_fragility}
\citet{mirzadeh2025} introduce \textsc{GSM-Symbolic}, which re-instantiates GSM8K \citep{cobbe2021training}, the foundational dataset of grade-school math word problems, from \emph{templates}: problem outlines whose entity names and numeric values are slots, so that the instances of one template share an outline and differ only in the slot values. They report that language model accuracy drops by up to 65\% when an irrelevant ``NoOp'' clause, a sentence that adds no information that a competent reasoner should use, is inserted into the prompt. They argue that this degradation reveals that LLMs solve math word problems by pattern matching rather than genuine reasoning. \citet{mirzadeh2025} is part of a broader line of work that probes math-reasoning robustness by injecting computationally irrelevant text into GSM8K problems, including \citet{shi2023distract}, \citet{li2024gsmplus}, and \citet{lai2025gsmdc}. These critiques agree at the behavioral level but stop short of localizing the failure inside the model's computation. Concurrently with our work, \citet{han2026fragile} give a mechanistic analysis of sensitivity to meaning-preserving perturbations and propose an architecture-level taxonomy of failure modes. They characterize how fragility differs across architectures, while we localize such failures internally. Concretely, we answer the question: when an LLM fails to solve a math word problem that contains an irrelevant ``NoOp'' clause, which modules inside the model are responsible?

\subsection{Mechanistic accounts of LLM math}
Earlier work provides accounts of how LLMs perform various skills involved in solving math word problems, including property representation \citep{gurnee2024}, entity binding \citep{wang2022ioi}, and arithmetic computation \citep{stolfo2023}. %
Each of these examines a particular skill required for solving math word problems, without analyzing how those skills compose into a complete solution. By contrast, we %
demonstrate that a four-stage reasoning pipeline organizes the full process in LLMs solving math word problems.

From the pipeline perspective, the work most closely related to ours is \citet{cheng2025can}, who identify a two-stage \emph{abstract-then-compute} mechanism for LLMs solving math word problems. They identify specific layers where abstraction occurs, followed by layers where arithmetic computation takes place. However, this two-stage pipeline is identified using synthetic one-step problems with restricted vocabulary, and is confined to the direct prompting regime, disabling Chain-of-Thought \citep{wei2022chain}, a standard component of LLM reasoning. Their analysis also focuses on the mechanisms underlying successful problem solving, rather than on model fragility. We close these gaps: we demonstrate that a staged reasoning pipeline operates on real, multi-step problems such as those in GSM8K, across direct and CoT regimes, and we identify the modules in this pipeline that are causally responsible for the failure mode.

\subsection{Mechanistic methods}
We use three standard tools from the mechanistic interpretability literature. Linear probing \citep{alain2017, conneau-etal-2018-cram} trains a linear regression on residual activations to test whether a target feature is linearly decodable. Since attention and MLP input projections and the unembedding all read the residual stream linearly \citep{elhage2021mathematical, elhage2022toy}, a probe drop shows that a feature has left the linearly accessible representation at the probed position. Activation patching \citep{vig2020, meng2022} replaces a target residual with a donor activation and measures the resulting change in output, identifying where causally relevant information lives. It, on the other hand, does not make any linearity assumption. Direct Logit Attribution \citep{wang2022ioi} projects each model component's output onto the unembedding direction of a target token, isolating which components write toward or %
away from that token.

\section{The Four-Stage Hypothesis}
\label{sec:four-stage}

\begin{figure*}[t]
  \centering
  \resizebox{0.98\textwidth}{!}{%
  \begin{tikzpicture}[
    xscale=1.333, font=\small,
    band/.style={draw=black!35, rounded corners=1.5pt, minimum height=6mm,
                 minimum width=2.4cm, inner sep=2pt, align=center,
                 font=\small\bfseries},
    state/.style={align=center, font=\small, text width=2.5cm, inner sep=1pt},
    rowlab/.style={anchor=east, font=\small, black!70},
    bnd/.style={font=\small, black!60, anchor=north},
    yes/.style={circle, fill=black!70, minimum size=1.9mm, inner sep=0pt},
    no/.style={draw=black!30, thin},
  ]
    \node[draw=black!35, rounded corners=3pt, fill=black!3, align=left,
          font=\small\itshape, text width=4.6cm, minimum height=2.42cm,
          inner sep=5pt] (prob) at (1.95,3.09)
      {``Janet's ducks lay 16 eggs a day. She eats 3 and bakes muffins with 4.
        She sells the rest at \$2 per egg. How much does she make daily?''};
    \node[font=\small\itshape, black!60, anchor=north] at (1.95,1.80) {input};
    \draw[-{Latex[length=1.8mm]}, black!55] (3.78,3.09) -- (4.22,3.09);

    \draw[draw=black!40, dashed, rounded corners=3pt] (4.20,1.70) rectangle (12.00,4.30);
    \node[font=\small\itshape, black!60, anchor=west] at (4.40,4.08)
      {layered transformation of the internal representation};

    \node[band, fill=blue!8]    at (5.20,3.42) {Schema\\Abstraction};
    \node[band, fill=orange!12, draw=red!55, thick] at (7.15,3.42) {Operation\\Planning};
    \node[band, fill=green!12]  at (9.10,3.42) {Operand\\Binding};
    \node[band, fill=purple!8]  at (11.05,3.42) {Computation};
    \foreach \a/\b in {6.10/6.25, 8.05/8.20, 10.00/10.15}
      \draw[-{Latex[length=1.6mm]}, black!45] (\a,3.42) -- (\b,3.42);

    \node[state] at (5.20,2.44) {narratives dropped relations kept};
    \node[state] at (7.15,2.56) {$R=(x{-}y{-}z)\cdot u$};
    \node[state] at (9.10,2.56) {$(16{-}3{-}4)\cdot 2$};
    \node[state] at (11.05,2.56) {$18$};
    \draw[red!55, thin] (7.15,2.32) -- (7.15,2.20);
    \node[font=\small, red!65, align=center, text width=3.2cm] at (7.15,2.03)
      {NoOp distractors corrupt this stage};

    \draw[black!40, thin] (4.30,1.62) -- (11.85,1.62);
    \foreach \x/\l in {4.30/0, 6.18/22, 8.13/36, 10.08/40, 11.85/80}
      {\draw[black!40, thin] (\x,1.66) -- (\x,1.58); \node[bnd] at (\x,1.54) {L\l};}

    \node[rowlab] at (4.10,1.06) {residual cosine};
    \node[rowlab] at (4.10,0.71) {linear probes};
    \node[rowlab] at (4.10,0.36) {activation patching};
    \node[rowlab] at (4.10,0.01) {head attribution};
    \foreach \x in {5.20, 7.15, 9.10, 11.05} \node[yes] at (\x,1.06) {};
    \foreach \x in {5.20, 9.10, 11.05}       \node[yes] at (\x,0.71) {};
    \draw[no] (7.06,0.71) -- (7.24,0.71);
    \foreach \x in {7.15, 9.10}              \node[yes] at (\x,0.36) {};
    \draw[no] (5.11,0.36) -- (5.29,0.36);
    \draw[no] (10.96,0.36) -- (11.14,0.36);
    \node[yes] at (7.15,0.01) {};
    \draw[no] (5.11,0.01) -- (5.29,0.01);
    \draw[no] (9.01,0.01) -- (9.19,0.01);
    \draw[no] (10.96,0.01) -- (11.14,0.01);
  \end{tikzpicture}}
  \caption{Overview of the four-stage pipeline, illustrated on one GSM8K problem.
  The problem is the model's input. The four stages are internal, successive
  transformations of the residual that represents the problem. Each stage leaves a distinct
  intermediate representation: surface narratives are stripped (Schema Abstraction), an operation plan is
  formed over abstracted variables (Operation Planning), prompt values are bound into it (Operand Binding), and the result is
  computed (Computation). Layer boundaries for Llama-3.3-70B are marked below, and the lower
  grid shows which method establishes each stage (filled dot) or does not speak
  to it (dash). Boundaries are placed by an automatic criterion on the dynamics of residual similarity between problems, and linear probes and activation patching are used to interpret the content of the representations at those boundaries
  (\S\ref{sec:four-stage-exp}). Direct logit attribution additionally localizes NoOp
  fragility to Operation Planning (\S\ref{sec:noop}).}
  \label{fig:framework}
\end{figure*}
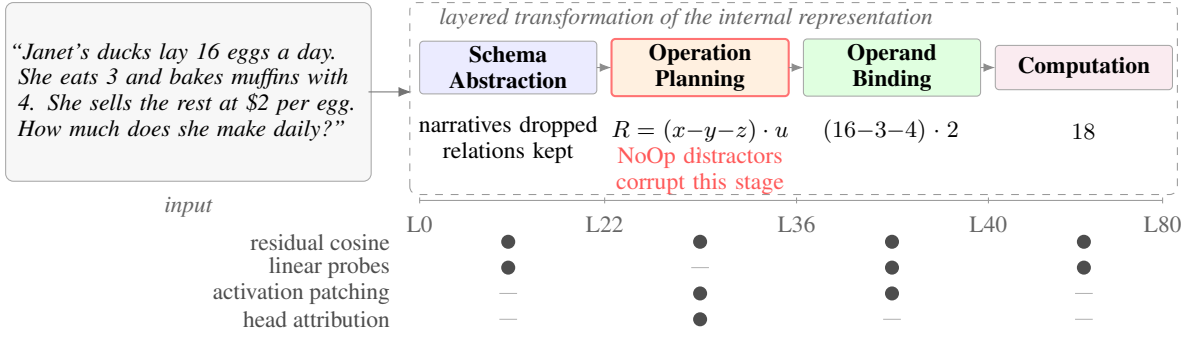

Refining on the two-stage abstract-then-compute view of \citet{cheng2025can}, we hypothesize that math word problem reasoning is composed of four stages. Each stage produces a distinct intermediate representation, and each can fail independently. Errors from any stage propagate and result in wrong final answers. Our decomposition is informed by cognitive science accounts of human math word-problem solving, which typically distinguish three or four cognitive operations: parsing the narrative into a propositional ``situation model'', selecting an appropriate solution plan, and executing the resulting computation \citep{kintschgreeno1985, hegartymayermonk1995}. 

Consider the GSM8K question: \emph{``Janet's ducks lay 16 eggs per day. She eats three for breakfast every morning and bakes muffins for her friends every day with four. She sells the remainder at the farmers' market daily for \$2 per fresh duck egg. How much in dollars does she make every day at the farmers' market?''} Solving this problem requires four distinct cognitive steps, each producing a result that the next step operates on. \textbf{Stage~1: Schema Abstraction.} The solver abstracts away surface narrative details such as entity names and story-specific wording (``Janet'', ``ducks'', etc.), while preserving the underlying relational structure of the problem: a quantity $x$ is produced daily, quantities $y$ and $z$ are consumed, the remainder is sold at price $u$ per unit, and the goal is revenue $R$. The resulting representation captures the problem schema, but not yet the operations.
\textbf{Stage~2: Operation Planning.} Identify the sequence of operations needed 
to solve the abstracted problem schema: $
R = (x - y - z) \cdot u$.  The output encodes the operation plan, but not yet the concrete operand values.
\textbf{Stage~3: Operand Binding.} Substitute concrete prompt values into the operation plan: $x\!\leftarrow\!16$ (eggs laid per day), $y\!\leftarrow\!3$ (eaten for breakfast), $z\!\leftarrow\!4$ (used for muffins), $u\!\leftarrow\!2$ (dollars per egg). The output is the fully grounded numeric expression $(16 - 3 - 4) \cdot 2$. \textbf{Stage~4: Computation.} The solver computes $(16 - 3 - 4) \cdot 2 = 18$ and reads out the answer $18$.

We ask whether LLMs develop an analogous sequential decomposition that appears as distinguishable transformations in the residual stream, and if so, which stage fails under distraction. \S\ref{sec:four-stage-exp} shows that the model’s internal representations undergo a four-stage transformation during math word-problem solving. \S\ref{sec:noop} then identifies which stage breaks under an irrelevant ``NoOp'' distractor.

\section{Experimental Setup}
\label{sec:setup}
We use Llama-3.3-70B-Instruct \citep{grattafiori2024llama} (80 layers) as the main target, with Qwen-2.5-14B-Instruct \citep{qwen2025} and Gemma-2-9B-it \citep{gemmateam2024gemma2} replications to show generalization in App. \ref{app:generalization}. All prompts follow the GSM8K answer format: the model emits the marker \texttt{\#\#\#\#} immediately before the final numeric answer (full template in App.~\ref{app:impl}).

We use four primary datasets, two from \citet{mirzadeh2025}: \textbf{GSM-Symbolic}, templated variants of GSM8K math word problems, and \textbf{GSM-P1}, which extends each GSM-Symbolic problem with one extra clause that requires one additional computation step. \citet{mirzadeh2025}'s NoOp dataset is not released, so we re-create \textbf{GSM-NoOp} via a human-in-the-loop process described in App.~\ref{app:impl}. We similarly construct a control \textbf{GSM-Filler}, matched in token length to the NoOp. Filler questions are of the same structure and contain the same numeric tokens as NoOp questions, but the Filler clauses are neutral descriptions that are less distracting than the NoOp. We also created \textbf{padded-Symbolic}, a length-controlled substitute for GSM-Symbolic used in the Stage~2 patching experiments (\S\ref{sec:formulation}). Each GSM-Symbolic question is padded with a computation-free clause matched in length to the extra clause of its GSM-P1 counterpart, so that a (padded-Symbolic, P1) pair differs in the operations the problem requires while matching narrative, operand values, and token length. Without this control, any patching effect between Symbolic and P1 could be attributed to their length difference rather than to the extra operation. App.~\ref{app:impl} gives details on all datasets and their construction. Table~\ref{tab:dataset-perf-app} reports Llama-3.3-70B's performance on every dataset.

All metrics are aggregated at the template level. For each template we compute the mean over its instances, then report the mean across templates. Confidence intervals are $95\%$ template-bootstrap CIs (cluster bootstrap with templates as the resampling unit), reflecting variability across problem outlines rather than within-template instance noise.

\section{Identifying the Four-Stage Pipeline}
\label{sec:four-stage-exp}

\begin{figure*}[!htbp]
  \centering
  \begin{subfigure}[!htbp]{0.48\textwidth}
    \includegraphics[width=0.90\linewidth]{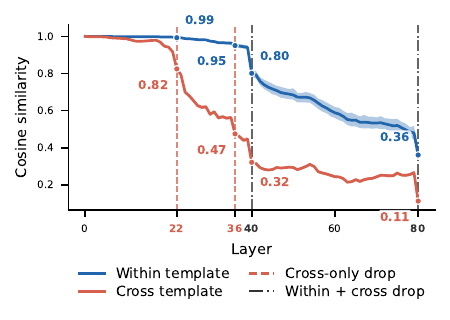}
  \end{subfigure}\hfill
  \begin{subfigure}[!htbp]{0.48\textwidth}
    \includegraphics[width=0.90\linewidth]{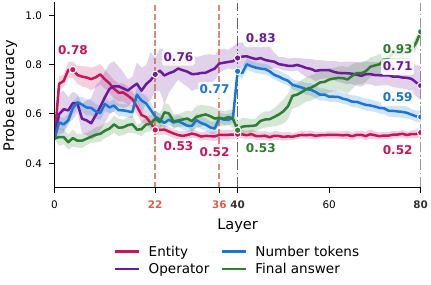}
  \end{subfigure}
  \caption{Four-stage signature in the answer-prefix position residual. \textbf{(a) Left: residual cosine similarity.} Mean pairwise cosine between within-template (blue) and cross-template (orange) prompts at each layer. Four drops emerge at \Labs{}, \Lform{}, \Lbind{}, and \Lout{}. \textbf{(b) Right: token-presence probes.} Per-layer linear-probe accuracy for entity tokens, operators, operand numbers, and the gold final-answer value. Entity decodability collapses at \Labs{} ($0.78 \to 0.53$), operand numbers jump to $0.77$ at \Lbind{}, and final-answer rises from chance to $0.93$ between \Lbind{} and \Lout{}. Robustness checks in App.~\ref{app:mode-robustness} and App~\ref{app:generalization}.}
  \label{fig:abstraction-main}
\end{figure*}

We extract the answer-prefix residual  %
(the residual stream at the token position immediately before the model emits the final answer)
 at every layer of the model on the GSM-Symbolic subset the model solves correctly ($n=1{,}200$ across $75$ templates under direct prompting). Two independent analyses on this layer sequence converge on the same boundary layers (Fig.~\ref{fig:abstraction-main}). A pairwise-similarity analysis (panel a) tracks \emph{when} the residual representation changes from layer to layer, and per-category linear probes (panel b) identify \emph{what} content enters or leaves at each transition.

\textbf{Residual similarity (Fig.~\ref{fig:abstraction-main} a).} At each layer we compute the cosine similarity between every pair of answer-prefix residuals and split pairs by template membership. \emph{Within} pairs share a problem outline, and \emph{cross} pairs do not. Four boundary drops emerge automatically (layer-to-layer drop $\geq 0.025$ at $z \geq 3$). At \Labs{}, cross-template similarity drops while within-template similarity stays flat. The model has begun separating different problem outlines but has not yet differentiated instances within an outline. At \Lform{}, cross-template similarity drops again with the within-template curve still flat, indicating further consolidation of the outline-level structure. At \Lbind{}, within-template similarity finally drops, signaling that instance-specific content such as operand numbers has entered the residual. Both curves decline together at \Lout{} as the computation concludes and the answer is read out.

\textbf{Residual content probes (Fig.~\ref{fig:abstraction-main}b).} At the same answer-prefix position, we train per-layer binary linear probes for four target categories: surface \emph{entity} names from the question narrative (``Janet'', ``ducks'', etc.), \emph{operators} (\texttt{+}, \texttt{-}, \texttt{$\times$}, \texttt{$\div$}), \emph{number} tokens used as operands, and the gold \emph{final-answer} value. Probes use template-disjoint cross-validation so no template appears in both train and test, forcing the probe to generalize across problem outlines rather than memorize template-specific encodings. The probe trajectories align with the boundaries from panel~(a). Entity decodability collapses at \Labs{} from $0.78$ to chance ($0.53$) and stays at chance thereafter, while operator decodability rises and remains elevated. By \Labs{} the residual has stopped representing surface words and now carries an abstract operator schema. Number-token accuracy jumps from chance to $0.77$ at \Lbind{}, where within-template similarity also first drops, indicating that template-specific operand numbers have entered the answer-position residual. Final-answer accuracy climbs from chance at \Lbind{} to $0.93$ at \Lout{}, the signature of computation.

The similarity analysis tells us \emph{when} the residual changes, and the probes tell us \emph{what} changes. The two readouts agree on three of the four layer indices ($\Labs{}/\Lbind{}/\Lout{}$) despite using no shared parameters. On that basis we identify the three boundaries, each boundary identified automatically from the similarity dynamics, with three stages, each stage's interpretation supported by the probes readout. \Labs{} marks Schema Abstraction (surface entities no longer represented), \Lbind{} marks Operand Binding (operands enter the residual at the answer prefix position), and \Lout{} marks Computation (the final answer appears). The staged structure is also observed in other datasets and domains. The same experiments reproduce all four signatures on SVAMP \citep{patel2021svamp}, and recovers a staged structure on PhantomWiki \citep{gong2025phantomwiki} multi-hop question answering (App.~\ref{app:cross-dataset}).

Besides these three aforementioned boundaries, the similarity analysis also indicates that cross-template residuals experience a drop in similarities at \Lform{}. However, direct probes show no significant results there so we cannot directly interpret what has entered the residuals at \Lform{}. Unlike the other stages, Operation Planning does not show up in direct probes since full operation plan has no objective label to probe against: the annotated formula is one of many equivalent forms and the model may plan in another. The causal patching experiments that follows in \S\ref{sec:formulation} fill in this gap, identifying \Lform{} as the  Operation Planning boundary.
Because Schema Abstraction, Operand Binding, and Computation are directly visible in the probe signatures, we focus the targeted patching analyses below on Operation Planning. \S\ref{sec:binding} adds a mechanistic account of the Operand Binding stage.

\subsection{Stage 2: Operation Planning}
\label{sec:formulation}
\begin{figure*}[!htbp]
  \centering
  \includegraphics[width=\textwidth]{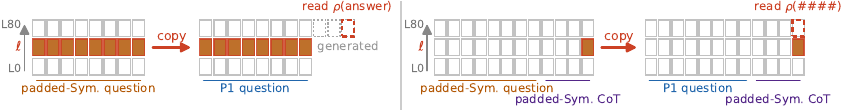}
  \caption{Operation Planning patching experiments. (padded-Symbolic, P1) pair differs in exactly one operation, so any causal effect is attributable to that
  operation alone.
  \textbf{Left:}~Question-span patching. Every question-span state is replaced, the
  model then generates its own CoT, and $\rho$ is read at the answer marker it
  emits, on the next-token $\log P(\text{P1's gold answer})$. An effect here means
  the operation plan has been formed in the patched activations.
  \textbf{Right:}~End-of-CoT patching. Only the final state of an injected CoT is
  replaced. That CoT solves padded-Symbolic but omits the step P1 requires, and
  $\rho$ is read at that last layer on the next-token
  $\log P(\texttt{\#\#\#\#})$. An effect here means the patched activations carry
  padded-Symbolic's plan, so the model treats the injected CoT as sufficient and
  commits.}
  \label{fig:patch-formulation-method}
\end{figure*}

We localize Operation Planning using activation patching between GSM-Symbolic and GSM-P1. For every Symbolic question, its P1 counterpart adds one extra clause that introduces one extra operation, so the two prompts differ in their required operation set. If the model plans operations during a stage, its representations of the two prompts must encode different plans by the end of that stage, and that difference must causally affect the final answer. We use padded-Symbolic as a length-controlled substitute for Symbolic (App.~\ref{app:impl} for details). We run two cross-prompt activation patching experiments on (padded-Symbolic, P1) pairs, restricted to instances where the model answers correctly on both sides ($n = 3{,}763$ across $87$ templates). Both setups are illustrated in Fig.~\ref{fig:patch-formulation-method}.

\begin{figure*}[!htbp]
  \centering
  \begin{subfigure}[!htbp]{0.48\textwidth}
    \includegraphics[width=0.90\linewidth]{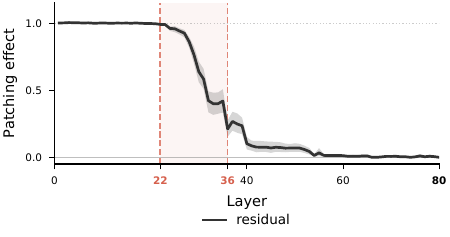}
  \end{subfigure}\hfill
  \begin{subfigure}[!htbp]{0.48\textwidth}
    \includegraphics[width=0.90\linewidth]{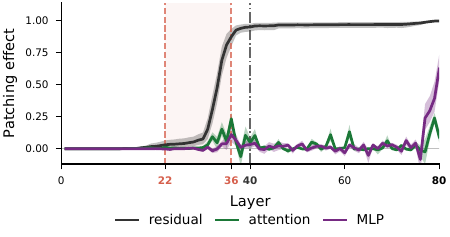}
  \end{subfigure}
  \caption{Stage~2: Operation Planning at \Lform{}. Both setups are defined in
    Fig.~\ref{fig:patch-formulation-method}. \textbf{(a) Left: CoT-mode question-span
    patching.} Patching effect stays near $1$ below \Labs{} and decays to $0.21$ by
    \Lform{}. By \Lform{}, the operation plan has been extracted and patching the question representations no longer changes the answer. \textbf{(b) Right: end-of-CoT patching}, decomposed by
    component. The full-residual trace rises from $0.08$ to $0.88$ across
    L30--\Lform{}. By \Lform{} the patched state at the end of CoT carries padded-Symbolic's
    plan, so the model treats the injected CoT as sufficient and commits.
    Attention spikes simultaneously with the rise for the full residual. \S\ref{sec:noop} further dissects the attention mechanism.}
  \label{fig:patch-formulation}
\end{figure*}

\textbf{(i) Question-span patching (Fig.~\ref{fig:patch-formulation}a).} At each layer, P1's residual at every problem-statement token position is overwritten by padded-Symbolic's residual at the same positions. The forward pass then resumes and the model freely generates its own CoT and final answer. We report the normalized patching effect $\rho$ on $\log P(\text{final answer})$ at the answer prefix. For each pair, $\rho = (\ell^{*}_{\text{tgt}} - \ell_{\text{tgt}}) / (\ell_{\text{src}} - \ell_{\text{tgt}})$, where $\ell_{\text{src}}$ and $\ell_{\text{tgt}}$ are the unpatched $\log P(\text{P1's gold answer})$ values on padded-Symbolic and P1 respectively, with $\rho = 0$ meaning no transfer and $\rho = 1$ meaning full transfer to padded-Symbolic's baseline log-probability. \textbf{(ii) Single-token patching at the end of an injected CoT (Fig.~\ref{fig:patch-formulation}b).} Both prompts are extended with the same CoT trace, taken from the gold CoT for the corresponding GSM-Symbolic instance. The trace solves padded-Symbolic correctly by construction but \emph{omits} the extra step P1 requires. At each layer we overwrite only P1's residual at the last token of the injected CoT (the position immediately before where the model would emit the answer prefix \texttt{\#\#\#\#}) and let the model continue. By giving P1 a CoT that is valid only for padded-Symbolic, we force the model to decide whether the provided trace matches the question's operation plan. If the plan has crystallized at layer $L$, patching padded-Symbolic's residual there should make the model commit immediately, raising $\log P(\texttt{\#\#\#\#})$ at that position. Otherwise the model continues to plan additional steps for P1's actual answer, and $\log P(\texttt{\#\#\#\#})$ remains low. We report the normalized patching effect $\rho$ on $\log P(\texttt{\#\#\#\#})$. The commit-readiness metric isolates the Operation Planning stage by testing whether the model has formed a plan consistent with the provided CoT, without
conflating that signal with later-stage answer production.
 
Both panels converge on the same transition boundaries. In Panel (a), padded-Symbolic's question-span residual fully restores P1's gold-answer log-probability at $\rho = 1.00$ before \Labs{}, then collapses across L22--\Lform{} to $\rho = 0.21$ at \Lform{}. By the end of Operation Planning, the model has extracted its operation plan from the input P1 question and no longer hands its answer over when patched. Panel (b) reports the complementary single-token result on commit-readiness. It rises from $0.08$ at L30 to $0.88$ at \Lform{}, saturating over the remaining layers. Decomposing the patching effects into the attention and MLP modules gives a first read on the underlying mechanism. The attention-output patch spikes at the same time as the full-residual rise, indicating that attention is the active component during Operation Planning. \S\ref{sec:noop} returns to this attention pathway with a finer head-level decomposition under NoOp-induced Operation Planning failure.

\subsection{Stage 3: Operand Binding}
\label{sec:binding}
GSM-Symbolic provides a natural testbed for localizing Operand Binding. Problems within a template share structure but differ in their instantiated operand values, so the representational gap between two same-template instances isolates exactly the binding of operands. We run two within-template patching experiments on length-matched pairs that the model solves correctly on both sides.

\begin{figure*}[t]
  \centering
  \begin{subfigure}[t]{0.48\textwidth}
    \includegraphics[width=0.90\linewidth]{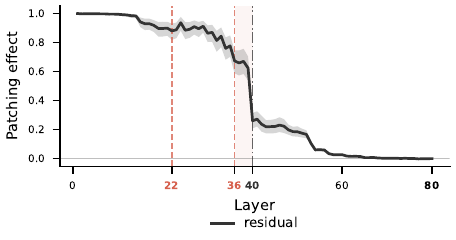}
  \end{subfigure}\hfill
  \begin{subfigure}[t]{0.48\textwidth}
    \includegraphics[width=0.90\linewidth]{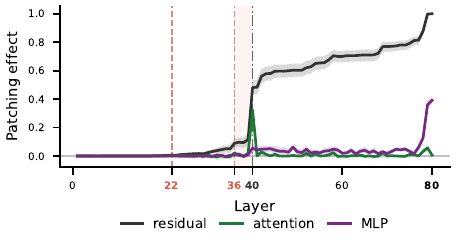}
  \end{subfigure}
  \caption{Stage 3: Operand Binding at \Lbind{}. Both panels patch between two
    instances of one template, which differ only in their operand values.
    \textbf{(a) Left: CoT-mode question-span patching.} The patching effect
    holds at $0.62$ at L39 and falls to $0.26$ at \Lbind{}, a single-layer
    collapse. By \Lbind{} the operands have been bound on the target side, so
    supplying the source's values no longer changes the answer.
    \textbf{(b) Right: direct-mode patching at the answer prefix}, decomposed
    by component. The full residual jumps from $0.12$ at L39 to $0.48$ at
    \Lbind{}, indicating that the operand values have entered the answer-prefix position. Patching the attention output alone reaches $0.32$ at \Lbind{}
    and returns to zero immediately after, while the MLP stays near zero until
    Computation. Attention is therefore what carries operand values into the
    answer-prefix residual, an IOI-style mover-head signature.}
  \label{fig:patch-binding}
\end{figure*}

\textbf{(i) Question-span patching (Fig.~\ref{fig:patch-binding}a).} At each layer, the target's residual at every question-token position is overwritten by the source's residual at the same positions. Two instances of one template differ only in their operand values, so replacing the question span replaces exactly those values and nothing else. The forward pass then resumes and the model freely generates its CoT. We report the normalized patching effect $\rho$ on $\log P(\text{target's gold answer})$ at the answer prefix, where $\rho = 0$ is no transfer and $\rho = 1$ is full transfer ($n = 1{,}387$ pairs across $99$ templates). \textbf{(ii) Single-token patching at the end of direct prompt (Fig.~\ref{fig:patch-binding}b).} In direct mode we patch the target's residual at only the answer-prefix position immediately before the model's emitted answer ($n = 333$ pairs across $50$ templates).

Both panels converge on \Lbind{} as the Operand Binding boundary. Panel (a) shows a sharp cliff at \Lbind{}: question-span transfer holds at $\rho = 0.62$ at L39 and falls to $0.26$ at \Lbind{}, a single-layer collapse. By \Lbind{} the operand binding has already been decided on the target side, so patching the question span after \Lbind{} can barely shift the answer. Panel (b) patches the answer-prefix position alone, decomposing the transport mechanism into attention and MLP modules. Attention spikes at \Lbind{}, while the MLP trace stays relatively flat with a late rise during Computation. This resembles
the IOI-style mover-head signature \citep{wang2022ioi}. At \Lbind{}, attention heads move operand information from the question's value-token positions into the answer-prefix residual, where it binds into the operation plan established upstream. 

Sections~\ref{sec:formulation} and~\ref{sec:binding} separate the Operation Planning and the Operand Binding stages using the same experiments run on different data pairs, which separates operation structure from operand values. The Operation Planning pair varies the required operations while matching operand values, and its effect concentrates in the band ending at \Lform{}. On the other hand, the Operand Binding pair varies operand values while holding the operation plan fixed, and its effect appears only at \Lbind{}.

\section{Diagnosing NoOp Fragility}
\label{sec:noop}
We localize NoOp fragility by mirroring the Stage~2 question-span patching from \S\ref{sec:formulation}, this time asking whether NoOp's wrong answer can be rescued by Filler's question representation. On (Filler-correct, NoOp-wrong) pairs ($n = 346$ across $25$ templates), at each layer we overwrite NoOp's residual at every problem-statement position with Filler's, let the model freely generate its CoT, and measure the patching effect on $\log P(\text{final answer})$, normalized so $1$ is full transfer and $0$ is none.

\begin{figure*}[!t]
  \centering
  \begin{subfigure}[t]{0.48\textwidth}
    \includegraphics[width=0.90\linewidth]{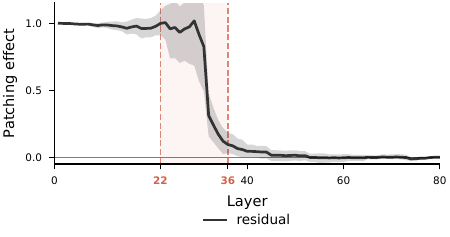}
  \end{subfigure}\hfill
  \begin{subfigure}[t]{0.48\textwidth}
    \includegraphics[width=0.90\linewidth]{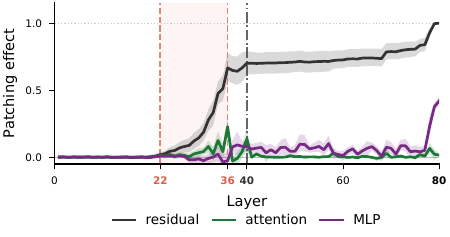}
  \end{subfigure}
  \caption{NoOp fragility localizes to Operation Planning. Both panels patch
    from Filler-correct into NoOp-wrong, a pair differing only in the distractor clause. \textbf{(a) Left: CoT-mode question-span patching.} The patching
    effect holds at $1.00$ through \Labs{} and collapses to $0.10$ by
    \Lform{}. The cliff falls in the same band as the clean Operation Planning
    transition (Fig.~\ref{fig:patch-formulation}), so what NoOp corrupts is
    the operation plan rather than an earlier or later stage.
    \textbf{(b) Right: direct-mode decomposition at the answer prefix},
    decomposed by component. The full residual rises from
    near zero at \Labs{} to $0.52$ at \Lform{}, localizing NoOp fragility to the Operation Planning band. Attention spikes simultaneously with the rise for the full residual, which we dissect further with the DLA.}
  \label{fig:noop-stage2}
\end{figure*}

Figure~\ref{fig:noop-stage2}a shows the patching effect staying near $1$ below \Labs{}, collapsing afterwards, and reaching to near $0$ by \Lform{}. The cliff sits in the same band as the clean Stage~2 transition (Fig.~\ref{fig:patch-formulation}), placing NoOp-induced failure inside Operation Planning: not a Schema-Abstraction error (which would corrupt before or at \Labs{}), not an Operand-Binding confusion (which would land at \Lbind{}), but a corrupted operation plan. Once that plan crystallizes at \Lform{}, the answer it commits to is locked in the residual stream and no longer responds to question-span patching. Figure~\ref{fig:noop-stage2}b decomposes the same effect in direct mode at the answer-prefix position. The full-residual trace rises across Stage~2 with the attention curve spiking in sync on and before \Lform{}, a first hint that attention carries the corruption.

\begin{figure}[tb]
  \centering
  \includegraphics[width=\columnwidth]{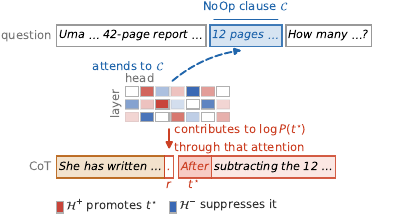}
  \caption{Engagement-anchored DLA. The target token $t^{\star}$ is the first token of the model-produced CoT
    that engages the distractor clause $\mathcal{C}$, shown in red. The readout
    position $r$ is the slot immediately before it. Eq.~\ref{eq:dla-clause} scores each
    head at each layer by the amount its write at $r$ adds to
    $\log P(t^{\star})$ through that head's own attention to $\mathcal{C}$. The score asks of each head, reading
    specifically the NoOp clause, whether it pushes the model toward or away from
    getting distracted by the clause.}
  \label{fig:dla-method}
\end{figure}

The cliff locates the corruption but does not name the heads that write it. To trace it, we use an \emph{engagement-anchored} per-head Direct Logit Attribution (DLA, illustrated in Fig.~\ref{fig:dla-method}). The anchor is the moment NoOp's own CoT first commits to the wrong plan: we find the first occurrence in the trace of a \emph{distractor-only} number (a number present in NoOp but absent from the matched Symbolic question), identify the sentence that contains it, and define the \emph{target token} $t^{\star}$ as the first token of that sentence, i.e.\ the opening of the sentence that first engages the distractor. The \emph{readout position} $r$ is the slot immediately before $t^{\star}$, where the model would emit it. The DLA then attributes each head's contribution to $\log P(t^{\star})$ at position $r$ that flows through attention to the NoOp clause. For head $h$ at layer $L$ and NoOp clause-token positions $\mathcal{C}$:
\begin{equation}
  \text{DLA}_h^{\mathcal{C}} \;=\; \sum_{p \in \mathcal{C}} \alpha_h[r, p] \cdot W_U[t^{\star}]^{\!\top}\, W_O^{h}\, V^{h}_p,
  \label{eq:dla-clause}
\end{equation}
where $\alpha_h[r,p]$ is the post-softmax attention weight that head $h$ attributes to clause token at position $p$ from readout position $r$, $V^{h}_p$ is head $h$'s value at $p$, $W_O^{h}$ is the head's slice of the output projection, and $W_U[t^{\star}]$ is the unembedding row for the per-example target token. Each summand factors into a \emph{routing} term (the attention weight $\alpha_h[r,p]$, i.e.\ how much $r$ attends to clause token $p$) and a \emph{writing} term (the OV product $W_U[t^{\star}]^{\!\top} W_O^{h} V^{h}_p$, the signed per-unit-attention contribution to $\log P(t^{\star})$). Summed over $\mathcal{C}$, the DLA isolates the head's contribution to $t^{\star}$ that flows specifically through attention to the NoOp clause.

We compute DLA on the NoOp-wrong subset whose CoT actively engages the distractor number ($n = 720$ across $37$ templates), and define two head sets within the Operation Planning band $L \in [23, 36]$: the \emph{anti-engagement} set $\mathcal{H}^{-} = \{(L,h) : \text{DLA}_h^{\mathcal{C},\text{noop}} < -10^{-4}\}$, whose clause attention writes \emph{against} $t^{\star}$ (suppressing the wrong-plan token), and the \emph{pro-engagement} set $\mathcal{H}^{+} = \{(L,h) : \text{DLA}_h^{\mathcal{C},\text{noop}} > +10^{-4}\}$, which writes \emph{for} it. The highest single-layer concentrations of both sets are found at \Lform{}, the Operation Planning boundary (App.~\ref{app:dla-full-layer}, Fig.~\ref{fig:dla-active-count-appendix}).

We test causal responsibility by scaling each head's output by a constant $\alpha$ throughout the forward pass and the generation rollout, for every head in sets $\mathcal{H}^{-}$, $\mathcal{H}^{+}$, and $\mathcal{H} = \mathcal{H}^{-} \cup \mathcal{H}^{+}$ respectively, and size-matched random heads drawn from layers \emph{outside} L23--L36. $\alpha = 1$ recovers the unperturbed model, $\alpha = 0$ ablates the head, and $\alpha = 2$ doubles its contribution. Two complementary experiments probe the heads' role in Operation Planning: \textbf{(i) Ablation damage on correct examples.} If $\mathcal{H}$ is the operation-planning machinery, ablating it on prompts the model normally solves should break the model. We ablate each head set at $\alpha = 0$ on the correctly-solved Symbolic subset ($n=4{,}752$, $100$ templates). %
\textbf{(ii) Amplification recovery on NoOp-wrong.} If $\mathcal{H}^{-}$ specifically suppresses the wrong-plan token $t^{\star}$ but is too weak to flip the answer on its own, amplifying it should rescue some of the lost accuracy. We amplify each head set at $\alpha = 2$ on the NoOp problems where the model answered wrongly and measure the fraction that recover the correct final answer.

Tables~\ref{tab:scaling-damage} and~\ref{tab:scaling-recovery} report the outcomes. Ablation collapses previously-correct accuracy by orders of magnitude on Symbolic-correct: $\mathcal{H}^{+}$ alone drops it to $0.06$, $\mathcal{H}^{-}$ alone to $0.47$, and their union to $0.02$, all against random-outside controls that retain $\geq 0.98$, with damage gaps reaching $\sim$10--20$\sigma$. Amplifying $\mathcal{H}^{-}$ on NoOp-wrong lifts accuracy from $0$ to $0.51$, $\sim 4\sigma$ over the $0.19$ random-outside baseline. Amplifying $\mathcal{H}^{+}$ lifts it to $0.33$ ($\sim 2\sigma$), consistent with these heads' Operation Planning role.

Together, the two interventions identify $\mathcal{H}$ as the causal instrument of Operation Planning. $\mathcal{H}^{+}$ is the dominant subset. Its ablation alone damages clean reasoning $7$--$8\times$ more than $\mathcal{H}^{-}$'s. When faced with NoOp distractors, the anti-engagement signal from $\mathcal{H}^{-}$ %
is overwhelmed by $\mathcal{H}^{+}$'s larger pro-engagement force. However, the DLA should not be read as a verdict on which heads cause the failure because Eq.~\ref{eq:dla-clause} sums only a head's direct write to $t^{\star}$ through its attention to the NoOp clause. $\alpha$-scaling, on the other hand, rescales the whole head over the full generation. Consequently, amplifying $\mathcal{H}^{+}$ helps as well. Moreover, the blunt $0.51$ recovery is not a repair ceiling. A single global scalar on a fixed head set recovers half of the failures, which places the failure at these heads and shows that a training-free lever exists there.

\begin{table}[tb]
\centering
\small
\begin{tabular}{lrr}
\toprule
Head set & $|\mathcal{H}|$ & Symbolic \\
\midrule
$\mathcal{H}^{-}$              & 134 & 0.47 $\pm$ 0.04 \\
$\mathcal{H}^{+}$             & 163 & 0.06 $\pm$ 0.01 \\
$\mathcal{H}$   & 297 & 0.02 $\pm$ 0.01 \\
\midrule
Random-135               & 135 & 0.98 $\pm$ 0.01 \\
Random-297               & 297 & 0.98 $\pm$ 0.01 \\
\bottomrule
\end{tabular}
\caption{Answer accuracy on Symbolic-correct after ablating each head set ($\alpha = 0$). Accuracy is $1$ before ablation by construction. $\mathcal{H}$ heads collapse accuracy by orders of magnitude relative to baseline.}
\label{tab:scaling-damage}
\end{table}

\begin{table}[tb]
\centering
\small
\begin{tabular}{lrr}
\toprule
Head set & $|\mathcal{H}|$ & NoOp \\
\midrule
$\mathcal{H}^{-}$              & 134 & \textbf{0.51 $\pm$ 0.06} \\
$\mathcal{H}^{+}$              & 163 & 0.33 $\pm$ 0.06 \\
$\mathcal{H}$   & 297 & 0.45 $\pm$ 0.07 \\
\midrule
Random-135               & 135 & 0.19 $\pm$ 0.05 \\
\bottomrule
\end{tabular}
\caption{Answer accuracy on NoOp-wrong after amplification ($\alpha = 2$). Accuracy is $0$ before amplification by construction. $\mathcal{H}^{-}$ gives a $+0.32$ gap over random-outside ($\sim 4\sigma$). $\mathcal{H}^{+}$ gives $+0.14$ ($\sim 2\sigma$).}
\label{tab:scaling-recovery}
\end{table}

\subsection{Classifying the Induced Failures}
\label{sec:error-typing}
The scaling experiments show that the heads are necessary for solving the problems. They do not by themselves show which stage breaks when they are absent. We therefore classify the failures induced when the heads are ablated.

On GSM-Symbolic problems the model normally solves correctly, we ablate a head set and read the resulting reasoning traces. We sample $60$ failure traces per head set and classify which stage the failures belong to. An Operation Planning failure reads the problem correctly but applies wrong or missing operations. Each trace is labeled independently by two annotators who see only the problem and the model's trace, and not which heads were turned off, so that the labels cannot be steered toward our hypothesis. A trace counts as a Planning failure only when both annotators agree.

\begin{table}[tb]
  \centering
  \small
  \begin{tabular}{lc}
    \toprule
    Head set & Operation Planning failures \\
    \midrule
    $\mathcal{H}$ & \textbf{0.91} \\
    Random-297      & 0.10 \\
    \bottomrule
  \end{tabular}
  \caption{Fraction of induced failures that are Operation Planning failures. Ablating the planning heads makes the model apply the wrong operations. Ablating size-matched random heads outside the band produces misreadings and arithmetic errors instead.}
  \label{tab:error-typing}
\end{table}

Table~\ref{tab:error-typing} reports the result. Ablating $\mathcal{H}$ produces Operation Planning failures in $0.91$ of cases, against $0.10$ for size-matched random heads outside the band, whose failures are misreadings and arithmetic slips instead. App.~\ref{app:error-typing} demonstrates the contrast through an example problem. The stage label is thus validated by what breaks: if \Lform{} implemented some other function, ablating its heads would produce that other failure type.

\section{Conclusion}
\label{sec:conclusion}
We presented a four-stage pipeline, Schema Abstraction, Operation Planning, Operand Binding, and Computation, that decomposes how LLMs process GSM-style math word problems. Residual cosine dynamics, presence probes, and cross-prompt activation patching converge independently on the same stage boundaries. The framework refines \citet{cheng2025can}'s two-stage abstract-then-compute account into four stages and extends it from synthetic one-step direct-mode prompts to multi-step CoT reasoning. Within this scaffold, NoOp fragility localizes to Operation Planning,
associated with a compact set of attention heads whose bidirectional causal role we validated by scaling: ablation collapses correct solutions by orders of magnitude against size-matched random controls, while amplifying the anti-engagement subset alone recovers a substantial fraction of NoOp failures. 
Our results suggest that distractor-induced failures arise from localized corruption of an intermediate Operation Planning stage rather than a global breakdown of reasoning.

\section*{Limitations}
Our analysis focuses on math word-problem reasoning and on distractor-induced failures in GSM-style tasks. The transfer results of \S\ref{sec:four-stage-exp} and App.~\ref{app:cross-dataset} are first steps beyond that scope, but they are probe-level replications rather than the full battery of methods applied in the main text, and mapping other tasks' pipelines at that depth, with probes, patching, and causal interventions, remains open. A second direction is repair. The amplification result in \S\ref{sec:noop} identifies the location for repair, and turning it into a targeted, deployable intervention is future work that the localization makes concrete.

\section*{Acknowledgments}
This project is partially supported by  an AI2050 Senior Fellowship, a Schmidt
Sciences program;
the National Science Foundation (NSF); the National Institute of Food and Agriculture (USDA/NIFA) (2023-67021-39829); the Air Force Office of Scientific Research (AFOSR) (FA9550-23-1-0322)

\bibliography{custom}

\clearpage
\appendix

\section{Implementation Details}
\label{app:impl}

\paragraph{Prompt template.} All experiments use zero-shot prompting with the model's native chat template. The system prompt is fixed across conditions:
\begin{quote}\small\ttfamily
You are a chatbot who is capable of performing the arithmetic problems.
\end{quote}
The user message concatenates the question with a mode-specific instruction.\\
For direct prompting:
\begin{quote}\small\ttfamily
\{question\}\\
Please answer the question directly WITHOUT showing the reasoning process. At the end, you MUST write the answer as an integer after `\#\#\#\#', without the equation or units.
\end{quote}
For chain-of-thought (CoT) prompting:
\begin{quote}\small\ttfamily
\{question\}\\
Let's think step by step. At the end, you MUST write the answer as an integer after `\#\#\#\#', without the equation or units.
\end{quote}
In CoT experiments that inject a cached reasoning trace (e.g., the Stage~2 single-token patching setup in \S\ref{sec:formulation}), the cached reasoning text, anything the model generates before the answer prefix \texttt{\#\#\#\#}, is inserted after the assistant header.

\paragraph{GSM-NoOp construction.} For each GSM-Symbolic template, Claude Opus 4.7 is prompted to propose three candidate NoOp clauses that read as natural continuations of the problem narrative. One is manually selected under two criteria: the clause must be inconsequential to the gold computation, and it must be plausible enough to function as a distractor. The selected clause is then programmatically instantiated across the template's entities and numbers (e.g.\ substituting matching characters, units, and quantities), producing one NoOp instance per Symbolic instance with one-to-one alignment. We sample a random subset of the resulting NoOp questions and manually verify that each reads naturally end-to-end, the inserted clause does not change the gold answer, and the clause is not trivially ignorable.

\paragraph{GSM-Filler construction.} GSM-Filler is built in parallel to NoOp using the same template-by-template pipeline. For each NoOp clause we author a filler clause that reuses the same numeric token(s) that appear in the matched NoOp clause but embeds them in a context with no computational implication (e.g., a descriptive or stylistic mention). Holding the surface number constant isolates the contribution of the surrounding semantics: any accuracy gap between GSM-Filler and GSM-NoOp on length-matched, digit-matched prompts attributes to how the model interprets the clause, not to which numeric tokens are present. Filler is then length-matched to NoOp at the prompt level by adding filler words like ``indeed'', ``as noted'' to the shorter side.

\paragraph{GSM-Filler-DF construction.} GSM-Filler-DF is a stricter digit-free baseline built with the same pipeline as GSM-Filler but with all number tokens stripped from the inserted clause and replaced by words like ``some'', i.e., words when substituting for a number token still make the question reads naturally. It rules out residual ``soft-NoOp'' effects that the digit-matched GSM-Filler may inherit from the surface number itself.

\paragraph{padded-Symbolic construction.} padded-Symbolic uses the same authoring pipeline as GSM-Filler-DF, with the filler clause length-matched to the corresponding GSM-P1 clause rather than to the NoOp clause. It serves as a length-controlled substitute for GSM-Symbolic in the Stage~2 patching experiments (\S\ref{sec:formulation}).

\paragraph{Dataset accuracy gradient.} Table~\ref{tab:dataset-perf-app} reports accuracy on all six datasets, including padded-Symbolic and the digit-free GSM-Filler-DF. Padded-Symbolic accuracy matches GSM-Symbolic to within $-0.004$ direct / $-0.016$ CoT, supporting its use as a length-controlled substitute in \S\ref{sec:formulation}. The digit-bearing GSM-Filler used in the main text shows a small soft-NoOp drop ($-0.067$ direct, $-0.052$ CoT) that is $\sim 2\times$ smaller than GSM-NoOp's in CoT and $\sim 1.5\times$ smaller in direct; the digit-free GSM-Filler-DF is statistically indistinguishable from baseline.

\begin{table}[tb]
  \centering
  \small
  \setlength{\tabcolsep}{4pt}
  \begin{tabular}{lrrrrr}
    \toprule
     & & \multicolumn{2}{c}{Direct} & \multicolumn{2}{c}{CoT} \\
    \cmidrule(lr){3-4} \cmidrule(lr){5-6}
    Dataset & $n$ & Acc. & $\Delta$ & Acc. & $\Delta$ \\
    \midrule
    GSM-Symbolic    & 5{,}000 & 0.241 & ---       & 0.950 & ---       \\
    padded-Symbolic & 4{,}852 & 0.237 & $-0.004$  & 0.934 & $-0.016$  \\
    GSM-P1          & 5{,}000 & 0.068 & $-0.173$  & 0.912 & $-0.038$  \\
    GSM-Filler-DF      & 5{,}000 & 0.237 & $-0.004$  & 0.941 & $-0.010$  \\
    GSM-Filler    & 5{,}000 & 0.174 & $-0.067$  & 0.898 & $-0.052$  \\
    GSM-NoOp        & 5{,}000 & 0.140 & $-0.101$  & 0.780 & $-0.171$  \\
    \bottomrule
  \end{tabular}
  \caption{Llama-3.3-70B accuracy across all datasets (50 templates $\times$ 100 instantiations each), including padded-Symbolic (the length-matched substitute used in the Stage~2 patching experiments) and the digit-free GSM-Filler-DF row. $\Delta$ is relative to GSM-Symbolic. Padded-Symbolic accuracy matches GSM-Symbolic to within $-0.004$ direct and $-0.016$ CoT, supporting its use as a length-controlled substitute in \S\ref{sec:formulation}. The digit-bearing GSM-Filler used in the main text yields a small but consistent ``soft-NoOp'' drop, strictly larger than the digit-free GSM-Filler-DF's null effect and strictly smaller than GSM-NoOp's drop.}
  \label{tab:dataset-perf-app}
\end{table}

\section{Prompting Mode Robustness}
\label{app:mode-robustness}
\label{app:presence-pre-reasoning} %

\S\ref{sec:four-stage-exp} uses the answer-position residual under direct prompting as the canonical readout for Figure~\ref{fig:abstraction-main}: direct prompting forces all internal computation through a single position with no externalized trace, cleanly exposing the staged structure. We verify here that the same signature appears under Chain-of-Thought (CoT) prompting and at a pre-reasoning readout where the model has not yet emitted any trace.

\begin{figure*}[!htbp]
  \centering
  \begin{subfigure}[!htbp]{0.48\textwidth}
    \includegraphics[width=\linewidth]{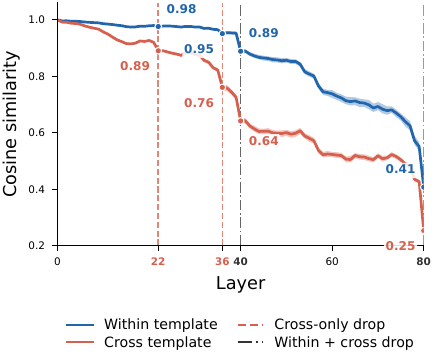}
  \end{subfigure}\hfill
  \begin{subfigure}[!htbp]{0.48\textwidth}
    \includegraphics[width=\linewidth]{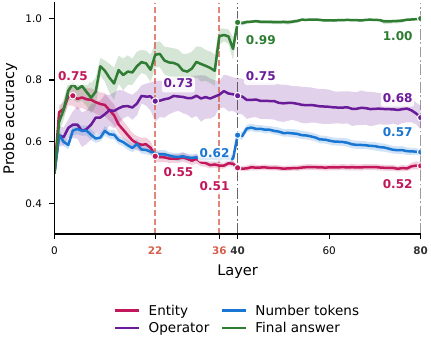}
  \end{subfigure}
  \caption{CoT-mode replication of Fig.~\ref{fig:abstraction-main}. \textbf{(a) Left: residual cosine similarity.} \textbf{(b) Right: token-presence probes.} The readout sits at the answer-prefix token \emph{following} the cached chain-of-thought trace, so attention at the readout position can attend over the trace tokens. The \Labs{} signature replicates in both panels: cross-template cosine drops, entity decodability collapses to chance, operator features rise. The \Lbind{} number-token jump is smaller than in direct mode (operands are attended-to via the trace) and the final-answer probe saturates early (the trace contains the answer string). Both are expected consequences of attention into the externalized reasoning, not contradictions of the staged internal structure exposed in Fig.~\ref{fig:abstraction-main}.}
  \label{fig:abstraction-cot}
\end{figure*}

\paragraph{CoT-mode replication.} Figure~\ref{fig:abstraction-cot} repeats Fig.~\ref{fig:abstraction-main} at the answer-prefix token \emph{following} the cached CoT trace. The \Labs{} signature replicates in both panels: cross-template cosine drops, entity decodability collapses to chance, and operator features rise. The Stage-3 and Stage-4 readouts shift as expected: the \Lbind{} number-token jump is smaller (operands are already attended to via the CoT trace) and the final-answer probe saturates early (the CoT trace text contains the answer already). These shifts are downstream consequences of attention into CoT trace tokens, not changes to the underlying stage structure.

\begin{figure}[tb]
  \centering
  \includegraphics[width=\columnwidth]{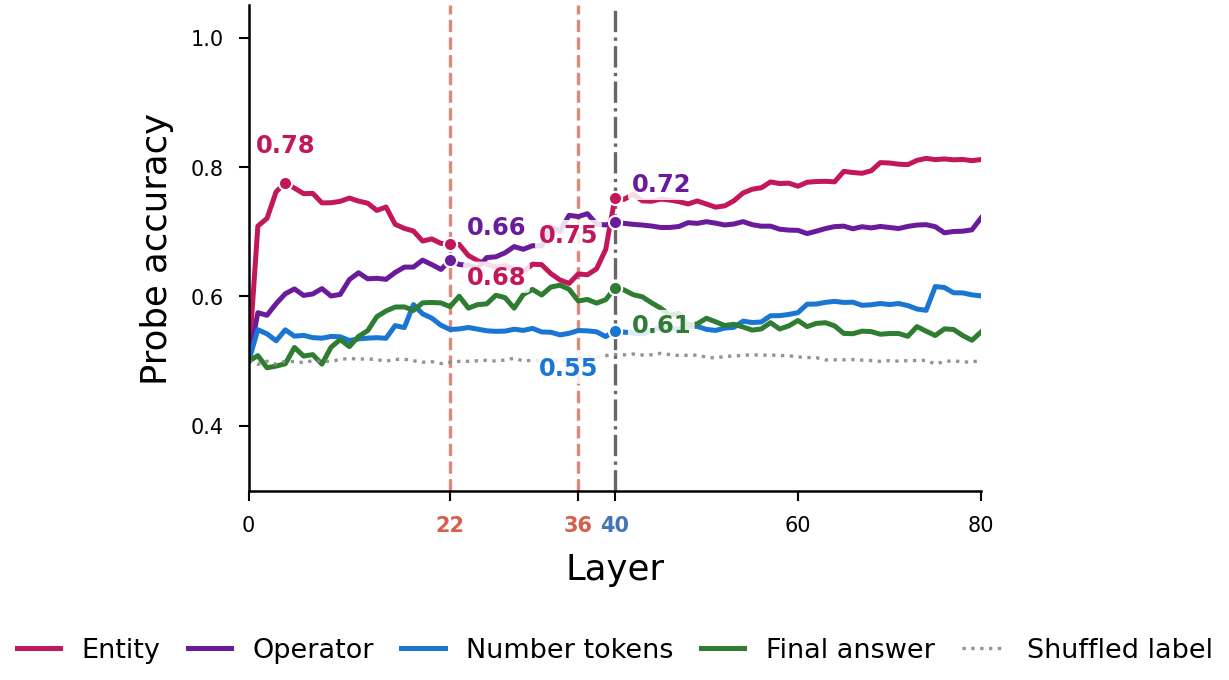}
  \caption{Linear presence probing at the assistant-header position of a CoT-framed prompt before any reasoning has been generated. The entity-drop / operator-rise dissociation at \Labs{} that defines Stage~1 in Fig.~\ref{fig:abstraction-main} reappears here. The drop is smaller than at the decision-point readouts in Fig.~\ref{fig:abstraction-main} and Fig.~\ref{fig:abstraction-cot}, consistent with stage circuitry engaging most strongly at output-commitment positions. Stage~3 (number-token) and Stage~4 (final-answer) signatures are absent at this position, consistent with binding and computation being done during CoT generation.}
  \label{fig:presence-pre-reasoning}
\end{figure}

\paragraph{Pre-reasoning matched-position control.} A complementary control reads at the assistant-header position of a CoT-framed prompt \emph{before any reasoning has been generated}: same prompt template as the CoT readout, but with no CoT trace tokens given as input. Figure~\ref{fig:presence-pre-reasoning} shows the \Labs{} entity-drop / operator-rise dissociation persists at this readout, attenuated but directionally identical (entity $0.78\to0.68$ here vs.\ $0.78\to0.53$ in direct mode). After dropping at \Labs{}, entity probe rises again at \Lbind{} since the entity names are relevant for writing the CoT trace. Number token and final answer accuracies stay close to chance throughout as the operand binding and computation happens throughout the CoT trace, instead of at the pre-CoT position. The three positions thus exhibit a coherent ordering (direct $\approx$ CoT $>$ pre-reasoning), consistent with \Labs{} circuitry engaging most strongly at decision-point tokens where the model is committing to an output.

\section{Filler-DF Replications}
\label{app:noop-stage2-fwd}

\begin{figure*}[!htbp]
  \centering
  \begin{subfigure}[!htbp]{0.48\textwidth}
    \includegraphics[width=\linewidth]{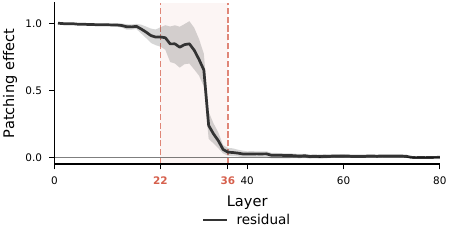}
  \end{subfigure}\hfill
  \begin{subfigure}[!htbp]{0.48\textwidth}
    \includegraphics[width=\linewidth]{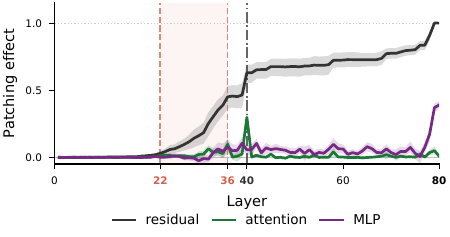}
  \end{subfigure}
  \caption{NoOp diagnosis of Llama-70B on the digit-free Filler-DF control, companion to Fig.~\ref{fig:noop-stage2}. \textbf{(a) Left: question-span CoT-mode patching.} From Filler-DF-correct to NoOp-wrong. The L22--\Lform{} cliff replicates. \textbf{(b) Right: direct-mode prompt-end decomposition.} Full residual (blue), attention output only (green), MLP output only (red). The full-residual rise replicates.}
  \label{fig:noop-stage2-fwd}
\end{figure*}

The main text presents the Stage~2 break test on the digit-bearing filler control (GSM-Filler). The length-matched digit-free variant (GSM-Filler-DF) has near-zero behavioral effect when compared with GSM-Symbolic (Table~\ref{tab:dataset-perf-app}), and serves as a different control that checks if the extra numeric token in the NoOp has any effect on model fragility. Figure~\ref{fig:noop-stage2-fwd} replicates both panels of the main-text Stage~2 break test (Fig.~\ref{fig:noop-stage2}) on the GSM-Filler-DF control ($n = 305$ across $19$ templates).

\paragraph{Question-span cliff (panel a).} The L22--\Lform{} cliff in the patching effect on $\log P(\text{final answer})$ replicates the digit-bearing variant in the main text, confirming the cliff is not specific to the digit-bearing filler.

\paragraph{Sublayer decomposition (panel b).} The answer-prefix sublayer decomposition (full residual / attention / MLP) for direct mode also replicates: the full-residual rise is present in the L22--\Lform{} band.

\section{Engagement-Anchored DLA: Full-Layer Scan}
\label{app:dla-full-layer}

\begin{figure}[!htbp]
  \centering
  \includegraphics[width=\columnwidth]{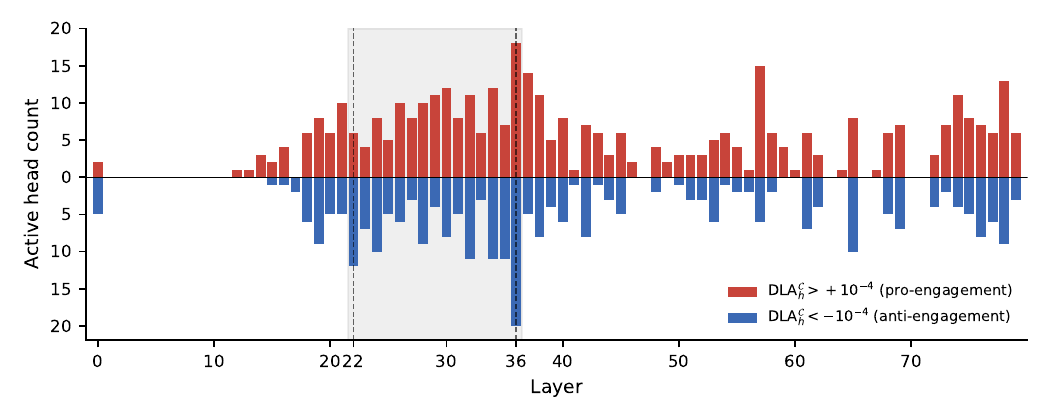}
  \caption{Per-layer count of heads with $|\mathrm{DLA}_h^{\mathcal{C},\,\mathrm{noop}}| > 10^{-4}$ across all 80 layers of Llama-3.3-70B. Bars above the zero-line: heads writing toward $t^{\star}$ (pro-engagement, $\mathcal{H}^{+}$). Bars below: heads writing against $t^{\star}$ (anti-engagement, $\mathcal{H}^{-}$). Gray shading marks the Operation Planning band. Activity ramps from near-zero pre-\Labs{}, reaches its highest density inside the band ($27.2\%$ of heads active vs.\ $5.5\%$ below \Labs{} and $12.9\%$ above \Lform{}), and peaks at \Lform{} with $20$ anti- + $18$ pro-engagement heads. Isolated post-\Lform{} layers (L57, L65, L74, L78) carry secondary activity, attributable to operand-binding and computation/readout machinery downstream of, and dependent on, operation planning.}
  \label{fig:dla-active-count-appendix}
\end{figure}

The main-text engagement-anchored DLA analysis is restricted to L22--\Lform{}, the operation-planning band identified independently by the patching results in \S\ref{sec:formulation}. To confirm that this band-restriction is not cherry-picking, we report results on the same DLA computation across all 80 layers of Llama-3.3-70B. Figure~\ref{fig:dla-active-count-appendix} shows the per-layer count of heads with $|\text{DLA}| > 10^{-4}$, split by anti- and pro-engagement direction. Three observations support the main-text scope:

\paragraph{Pre-\Labs{} negative control.} Layers L0--L22 are essentially silent: only $5.5\%$ of heads exceed $|\text{DLA}| > 10^{-4}$, with maximum $|\text{DLA}|$ below $0.0026$ at any head. The engagement-anchored measurement does not produce spurious signal in pre-planning layers.

\paragraph{Density peaks inside the Operation Planning band.} Inside the band, $27.2\%$ of heads exceed the active-head threshold, with mixed-sign structure (anti- and pro-\texttt{\#\#\#\#}) consistent with the main-text $\mathcal{H}^{-}$/$\mathcal{H}^{+}$ decomposition. This is the highest active-head density in the layer sweep.

\paragraph{Post-\Lform{} activity is real but downstream.} After \Lform{}, $12.9\%$ of heads remain above threshold, and several layers (L57, L65, L74, L78) carry isolated large-magnitude heads. We interpret these as the operand-binding attention mover (\S\ref{sec:binding}) and computation machinery downstream of, and dependent on, operation planning. 

\section{Failure Classification: A Worked Example}
\label{app:error-typing}

The following problem was solved correctly by the unablated model and appears in
both ablation conditions of \S\ref{sec:error-typing}, so the two failure types can
be compared on a single item. \emph{A fog bank takes $398$ minutes to cover every
$3$ miles of a $42$-mile city. How long will it take to cover the whole city?}
(gold answer $5{,}572$).

With size-matched random heads ablated, the model forms the correct plan and slips
only in the arithmetic: it computes the per-mile rate $398 \div 3$, multiplies by
$42$, and answers $5570$ instead of $5572$. Comprehension and operation selection
are intact, only the final value is wrong, which is not an Operation Planning failure.

With the $\mathcal{H}$ heads ablated, the model reads the same quantities and
still computes the per-mile rate correctly, then fails to form the multiplication.
It asserts that covering the whole city takes the same $398$ minutes as covering
$3$ miles, and answers $398$, a quantity copied from the question. Comprehension is
intact and no arithmetic is botched. The operation plan is never completed.

The blind annotation sheets and the adjudicated per-trace labels for all sampled
failures are included in the code release.

\section{Generalization to Other Models}
\label{app:generalization}

We replicate the Stage~1 signatures from Section~\ref{sec:four-stage-exp} on two additional open-source models: Gemma-2-9B-it (42 layers) and Qwen-2.5-14B-Instruct (48 layers). The four-stage residual-stream structure reproduces in both.

\begin{table}[!htbp]
  \centering
  \small
  \setlength{\tabcolsep}{3pt}
  \begin{tabular}{llrrrr}
    \toprule
     & & \multicolumn{2}{c}{Direct} & \multicolumn{2}{c}{CoT} \\
    \cmidrule(lr){3-4} \cmidrule(lr){5-6}
    Model & Dataset & Acc. & $\Delta$ & Acc. & $\Delta$ \\
    \midrule
    \multirow{5}{*}{Llama-3.3-70B}
      & Symbolic & 0.241 & ---      & 0.950 & ---      \\
      & P1       & 0.068 & $-0.173$ & 0.912 & $-0.038$ \\
      & Filler-DF& 0.237 & $-0.004$ & 0.941 & $-0.010$ \\
      & Filler   & 0.174 & $-0.067$ & 0.898 & $-0.052$ \\
      & NoOp     & 0.140 & $-0.101$ & 0.780 & $-0.171$ \\
    \midrule
    \multirow{5}{*}{Gemma-2-9B}
      & Symbolic & 0.125 & ---      & 0.859 & ---      \\
      & P1       & 0.036 & $-0.089$ & 0.761 & $-0.098$ \\
      & Filler-DF& 0.120 & $-0.005$ & 0.856 & $-0.003$ \\
      & Filler   & 0.079 & $-0.046$ & 0.783 & $-0.076$ \\
      & NoOp     & 0.062 & $-0.063$ & 0.556 & $-0.303$ \\
    \midrule
    \multirow{5}{*}{Qwen-2.5-14B}
      & Symbolic & 0.162 & ---      & 0.918 & ---      \\
      & P1       & 0.059 & $-0.103$ & 0.878 & $-0.040$ \\
      & Filler-DF& 0.159 & $-0.003$ & 0.913 & $-0.005$ \\
      & Filler   & 0.127 & $-0.035$ & 0.897 & $-0.021$ \\
      & NoOp     & 0.083 & $-0.079$ & 0.762 & $-0.156$ \\
    \bottomrule
  \end{tabular}
  \caption{Four-dataset accuracies for the three models (companion to Table~\ref{tab:dataset-perf-app}). $\Delta$ is relative to GSM-Symbolic for each model. The Filler-DF-null / Filler-small-drop / P1-drop / NoOp-largest-drop ordering replicates across all three model families, with digit-bearing Filler producing a small ``soft-NoOp'' drop strictly between digit-free Filler-DF and full NoOp.}
  \label{tab:dataset-perf-robust}
\end{table}
\paragraph{Dataset accuracies.} Table~\ref{tab:dataset-perf-robust} reports the four-dataset accuracies for Gemma-2-9B and Qwen-2.5-14B alongside Llama-3.3-70B (companion to Table~\ref{tab:dataset-perf-app}). The qualitative pattern from Llama replicates in both: Filler-DF is nearly indistinguishable from Symbolic ($\lvert\Delta\rvert \leq 0.005$), P1 incurs a moderate drop, and NoOp is the largest behavioral hit.

\begin{figure*}[!htbp]
  \centering
  \begin{subfigure}[t]{0.48\textwidth}
    \includegraphics[width=\linewidth]{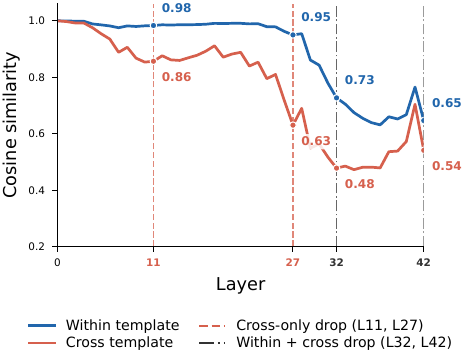}
  \end{subfigure}\hfill
  \begin{subfigure}[t]{0.48\textwidth}
    \includegraphics[width=\linewidth]{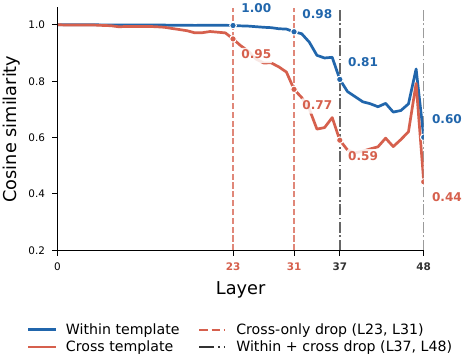}
  \end{subfigure}
  \caption{Direct-mode residual cosine similarity dynamics for two additional models (companion to Fig.~\ref{fig:abstraction-main}a). \textbf{(a) Left: Gemma-2-9B (42 layers).} \textbf{(b) Right: Qwen-2.5-14B (48 layers).} Each panel marks the model's own stage boundaries derived from the same automatic criterion used for Llama. Both models reproduce the four-event signature.}
  \label{fig:gen-template-sim}
\end{figure*}

\paragraph{Template similarity.} Figure~\ref{fig:gen-template-sim} shows residual cosine similarity dynamics for each model in direct mode with model-specific stage boundaries marked. Both Gemma-9B and Qwen-14B reproduce the canonical four-event sequence from Fig.~\ref{fig:abstraction-main}a: a first cross-only drop (Schema Abstraction), a second cross-only drop (Operation Planning), a joint within+cross drop (Operand Binding), and a terminal-layer readout (Computation).

\paragraph{Stage-boundary depth across models.} Table~\ref{tab:depth-ratios} reports each model's stage-boundary layers as fractions of total depth. Stage~1 (Schema Abstraction) completes at roughly the same proportional depth on Llama and Gemma ($0.26$--$0.28$) but considerably deeper on Qwen ($0.48$), so Qwen spends nearly half of its layers stripping surface narrative before the operation plan begins to form. Stage~2 (Operation Planning) and Stage~3 (Operand Binding) both shift later as a fraction of depth on Gemma and Qwen ($\sim 0.65$ and $\sim 0.77$) than on Llama ($0.45$ and $0.50$), leaving Llama with the longest tail of Stage~4 (Computation) layers in proportional terms ($50\%$ of the model vs.\ $23$--$24\%$ on the smaller models).

\begin{table}[tb]
  \centering
  \small
  \setlength{\tabcolsep}{5pt}
  \begin{tabular}{lccc}
    \toprule
    & Llama & Gemma & Qwen \\
    & 3.3-70B & 2-9B & 2.5-14B \\
    & (80L) & (42L) & (48L) \\
    \midrule
    \Labs{}  & 22 (0.28) & 11 (0.26) & 23 (0.48) \\
    \Lform{} & 36 (0.45) & 27 (0.64) & 31 (0.65) \\
    \Lbind{} & 40 (0.50) & 32 (0.76) & 37 (0.77) \\
    \Lout{}  & 80 (1.00) & 42 (1.00) & 48 (1.00) \\
    \bottomrule
  \end{tabular}
  \caption{Stage-boundary layers and their depth ratios (layer / total layers) for each model.}
  \label{tab:depth-ratios}
\end{table}

\begin{figure*}[!htbp]
  \centering
  \begin{subfigure}[!htbp]{0.48\textwidth}
    \includegraphics[width=\linewidth]{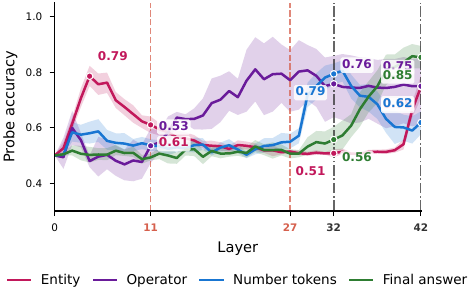}
  \end{subfigure}\hfill
  \begin{subfigure}[!htbp]{0.48\textwidth}
    \includegraphics[width=\linewidth]{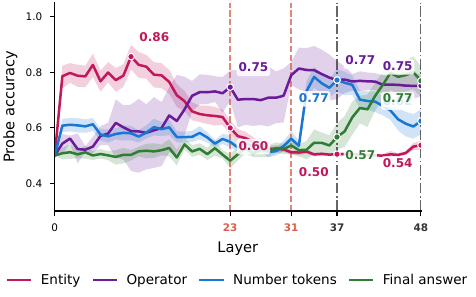}
  \end{subfigure}
  \caption{Linear presence-probe accuracies for the two additional models, companion to Fig.~\ref{fig:abstraction-main}b. \textbf{(a) Left: Gemma-2-9B (42 layers).} \textbf{(b) Right: Qwen-2.5-14B (48 layers).} Each panel marks the model's own stage boundaries. Entity decodability collapses at Stage 1 boundary, number-token decodability jumps at Stage 3 boundary, and final-answer accuracy rises from Stage 3 boundary through the end, matching the Llama signature and the boundaries identified independently from cosine dynamics in Fig.~\ref{fig:gen-template-sim}.}
  \label{fig:gen-presence-probe}
\end{figure*}

\paragraph{Presence probes.} Figure~\ref{fig:gen-presence-probe} replicates Fig.~\ref{fig:abstraction-main}b setup on Gemma-9B and Qwen-14B with each model's own stage boundaries marked. All three probed stage signatures reproduce: \textbf{(Stage~1)} entity decodability falls from a pre-cliff peak ($\sim 0.80$--$0.91$) to chance ($\sim 0.51$--$0.55$) by the cross-only boundary identified independently in Fig.~\ref{fig:gen-template-sim}; \textbf{(Stage~3)} number-token decodability jumps from chance to $\sim 0.78$--$0.81$ at the within-cross boundary; \textbf{(Stage~4)} final-answer decodability rises from chance to $\sim 0.78$--$0.86$ approaching the terminal layer.

\begin{figure*}[!htbp]
  \centering
  \begin{subfigure}[!htbp]{0.48\textwidth}
    \includegraphics[width=\linewidth]{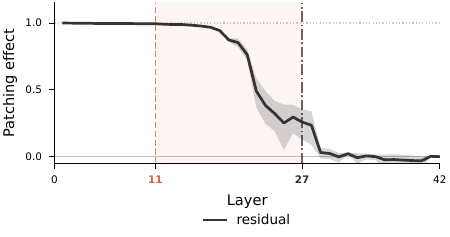}
  \end{subfigure}\hfill
  \begin{subfigure}[!htbp]{0.48\textwidth}
    \includegraphics[width=\linewidth]{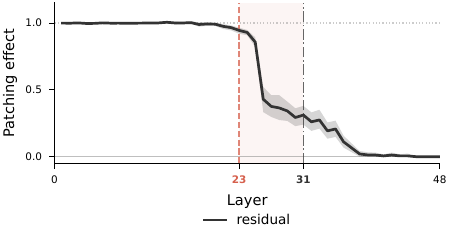}
  \end{subfigure}
  \caption{Stage~2 question-span patching (companion to Fig.~\ref{fig:patch-formulation}a) on the robustness models. \textbf{(a) Left: Gemma-2-9B.} \textbf{(b) Right: Qwen-2.5-14B.} Padded-Symbolic's question-span residual is patched into P1; we report the normalized patching effect on $\log P(\text{final answer})$. The trace rides near $1$ before each model's Schema Abstraction boundary and collapses by Operation Planning boundary, matching the Llama pattern.}
  \label{fig:gen-stage2-a}
\end{figure*}

\begin{figure*}[!htbp]
  \centering
  \begin{subfigure}[!htbp]{0.48\textwidth}
    \includegraphics[width=\linewidth]{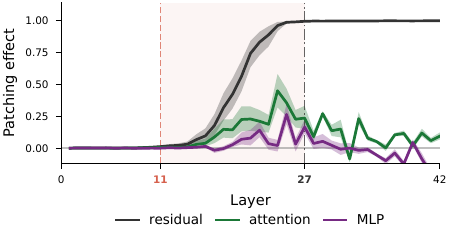}
  \end{subfigure}\hfill
  \begin{subfigure}[!htbp]{0.48\textwidth}
    \includegraphics[width=\linewidth]{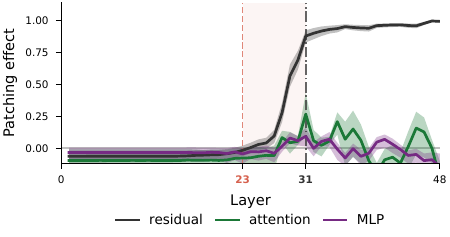}
  \end{subfigure}
  \caption{Stage~2 commit-readiness at the end of an injected GSM-Symbolic CoT (companion to Fig.~\ref{fig:patch-formulation}b), decomposed into full residual / attention / MLP. \textbf{(a) Left: Gemma-2-9B.} \textbf{(b) Right: Qwen-2.5-14B.} The full-residual trace rises across each model's Operation Planning band.}
  \label{fig:gen-stage2-c}
\end{figure*}

\paragraph{Stage 2 patching.} Figures~\ref{fig:gen-stage2-a} and~\ref{fig:gen-stage2-c} replicate the main-text Stage~2 panels (Fig.~\ref{fig:patch-formulation}) on Gemma-2-9B and Qwen-2.5-14B, with each model's own stage boundaries marked. The question-span patching effect on $\log P(\text{final answer})$ rides near $1$ before each model's Schema-Abstraction boundary (Gemma L11, Qwen L23) and collapses by the Operation-Planning boundary (Gemma L27, Qwen L31). For the single-token patch at the end of the injected CoT, the full-residual commit-readiness trace rises across the same band, with attention as the leading sublayer component, matching the Llama Stage~2 signature. Gemma has $n=864$ examples across $48$ templates. Qwen has $n=1{,}044$ across $36$ templates.

\begin{figure*}[!htbp]
  \centering
  \begin{subfigure}[!htbp]{0.48\textwidth}
    \includegraphics[width=\linewidth]{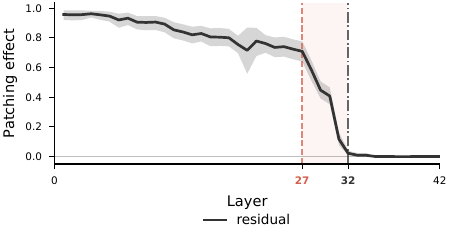}
  \end{subfigure}\hfill
  \begin{subfigure}[!htbp]{0.48\textwidth}
    \includegraphics[width=\linewidth]{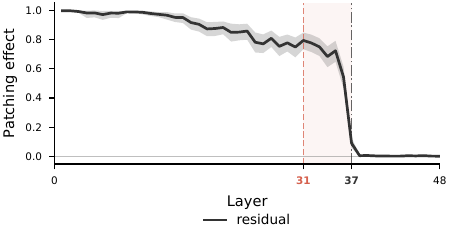}
  \end{subfigure}
  \caption{Stage~3 within-template value-token patching (companion to Fig.~\ref{fig:patch-binding}a) on the robustness models. \textbf{(a) Left: Gemma-2-9B.} \textbf{(b) Right: Qwen-2.5-14B.} Operand Binding cliff replicates inside each model's band.}
  \label{fig:gen-stage3-cot}
\end{figure*}

\begin{figure*}[!htbp]
  \centering
  \begin{subfigure}[t]{0.48\textwidth}
    \includegraphics[width=\linewidth]{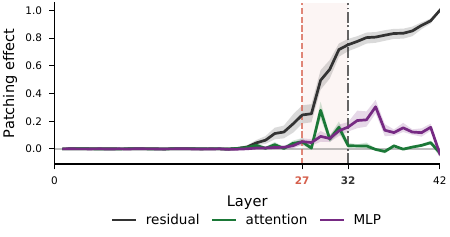}
  \end{subfigure}\hfill
  \begin{subfigure}[t]{0.48\textwidth}
    \includegraphics[width=\linewidth]{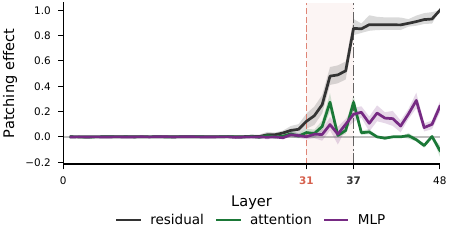}
  \end{subfigure}
  \caption{Stage~3 direct-mode patching at the answer prefix (companion to Fig.~\ref{fig:patch-binding}b), decomposed into full residual / attention / MLP. \textbf{(a) Left: Gemma-2-9B.} \textbf{(b) Right: Qwen-2.5-14B.} The residual curve raises and attention-output trace spikes inside each model's Operand Binding band.}
  \label{fig:gen-stage3-direct}
\end{figure*}

\paragraph{Stage 3 patching.} Figures~\ref{fig:gen-stage3-cot} and~\ref{fig:gen-stage3-direct} replicate the main-text Stage~3 panels (Fig.~\ref{fig:patch-binding}) on the two additional models, with each model's own stage boundaries marked. On these two models the CoT-mode patch is restricted to the numeric-value-token positions rather than applied to the whole question span as in Fig.~\ref{fig:patch-binding}a. Within a template the two are equivalent, since the source and target questions differ only in their operand values, and the value-token form additionally holds every operand at the same absolute position in both prompts. The CoT-mode within-template value-token cliff sits at the Operation-Planning $\to$ Operand-Binding transition (Gemma L27$\to$L32, Qwen L31$\to$L37). Gemma sample sizes are $n=926$ pairs across $87$ templates (CoT-mode) and $n=478$ across $47$ (direct-mode). Qwen has $n=711$ across $81$ templates (CoT-mode) and $n=617$ across $49$ (direct-mode).

\begin{figure*}[!htbp]
  \centering
  \begin{subfigure}[!htbp]{0.48\textwidth}
    \includegraphics[width=\linewidth]{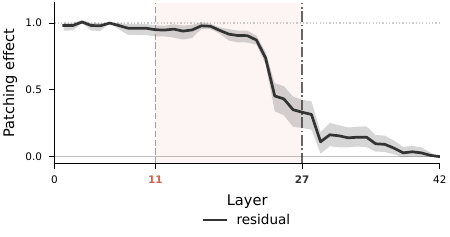}
  \end{subfigure}\hfill
  \begin{subfigure}[!htbp]{0.48\textwidth}
    \includegraphics[width=\linewidth]{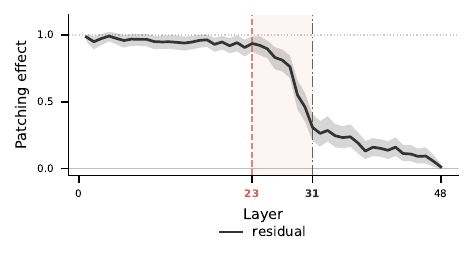}
  \end{subfigure}
  \caption{NoOp fragility, CoT-mode question-span patching, Filler-correct $\to$ NoOp-wrong (companion to Fig.~\ref{fig:noop-stage2}a). \textbf{(a) Left: Gemma-2-9B.} \textbf{(b) Right: Qwen-2.5-14B.} The Operation Planning stage cliff replicates on both models.}
  \label{fig:gen-noop-stage2-a}
\end{figure*}

\paragraph{NoOp fragility.} Figure~\ref{fig:gen-noop-stage2-a} replicates the main-text NoOp diagnosis (Fig.~\ref{fig:noop-stage2}a) on the Filler control for both models. The CoT-mode question-span cliff sits inside each model's Schema-Abstraction to Operation-Planning band (Gemma L11--L27, Qwen L23--L31). The per-layer sweep retains $738$ pairs across $44$ templates on Gemma and $140$ pairs across $43$ templates on Qwen.

\clearpage

\begin{table*}[t]
\centering
\small
\begin{tabular}{lrrr}
\toprule
Head set & $|\mathcal{H}|$ & Gemma & Qwen \\
\midrule
$\mathcal{H}^{-}$        & 74 / 122  & 0.27 $\pm$ 0.06 & \textbf{0.00 $\pm$ 0.00} \\
$\mathcal{H}^{+}$        & 134 / 118 & \textbf{0.00 $\pm$ 0.01} & 0.23 $\pm$ 0.07 \\
\midrule
Random-outside           & 134 / 118 & 0.82 $\pm$ 0.05 & 0.91 $\pm$ 0.03 \\
\bottomrule
\end{tabular}
\caption{Answer accuracy on Symbolic-correct after ablating each head set ($\alpha=0$); companion to Table~\ref{tab:scaling-damage} on Llama. Random-outside is size-matched to $|\mathcal{H}^{+}|$ for each model. Bold marks the subset whose ablation collapses accuracy to near-zero.}
\label{tab:scaling-damage-robust}
\end{table*}

\begin{table*}[t]
\centering
\small
\begin{tabular}{lrrr}
\toprule
Head set & $|\mathcal{H}|$ & Gemma & Qwen \\
\midrule
$\mathcal{H}^{-}$        & 74 / 122  & 0.31 $\pm$ 0.08 & \textbf{0.26 $\pm$ 0.09} \\
$\mathcal{H}^{+}$        & 134 / 118 & \textbf{0.37 $\pm$ 0.10} & 0.24 $\pm$ 0.09 \\
\midrule
Random-outside           & 74 / 122  & 0.22 $\pm$ 0.07 & 0.21 $\pm$ 0.08 \\
\bottomrule
\end{tabular}
\caption{Answer accuracy on NoOp-wrong runs after amplification ($\alpha=2$); companion to Table~\ref{tab:scaling-recovery} on Llama. Random-outside is size-matched to $|\mathcal{H}^{-}|$ for each model. Aggregates restricted to templates with $\geq 5$ NoOp-wrong pairs to avoid single-pair noise.}
\label{tab:scaling-recovery-robust}
\end{table*}

\paragraph{Head-scaling validation.} Tables~\ref{tab:scaling-damage-robust} and~\ref{tab:scaling-recovery-robust} replicate the Llama scaling experiments (Tables~\ref{tab:scaling-damage},~\ref{tab:scaling-recovery}) on Gemma-2-9B and Qwen-2.5-14B. Each model's $\mathcal{H}^{-}$ and $\mathcal{H}^{+}$ are identified by the same engagement-anchored DLA procedure, filtered to its own Operation Planning band. Random-outside controls are sampled from layers outside that band and size-matched to the corresponding head set. Recovery aggregates restrict to templates with $\geq 5$ NoOp-wrong pairs (template-cluster bootstrap is dominated by per-template variance, and the natural NoOp-wrong pool is heavily long-tailed: a sizeable fraction of templates produce only one such pair). \textbf{Ablation replicates on both models}: one head subset collapses Symbolic-correct accuracy by orders of magnitude relative to its size-matched random control ($\mathcal{H}^{+}$ for Gemma, $\mathcal{H}^{-}$ for Qwen). \textbf{Amplification recovery also replicates}: both head subsets recover NoOp-wrong examples above the random-outside baseline on both models, with Gemma's $\mathcal{H}^{+}$ showing the largest gap ($+0.15$) and Qwen's gaps smaller. Which subset acts as the dominant damage / rescue channel is therefore model-specific, but the existence of a compact Stage-2 head set whose ablation and amplification both move accuracy preferentially over random-outside controls is preserved across all three models.

\section{Generalization to Other Datasets and Tasks}
\label{app:cross-dataset}

The main text establishes the four-stage pipeline on GSM-Symbolic and its controlled variants. This appendix reports two transfer experiments, one to a different benchmark of math word problems and one to a non-math reasoning task. Both use the probe pipeline of \S\ref{sec:four-stage-exp} unchanged: the same answer-prefix readout, the same balanced-accuracy metric against an empirical shuffled-label baseline, and no per-benchmark tuning of any kind. Layer indices refer to Llama-3.3-70B throughout.
\paragraph{Cross-benchmark replication on SVAMP.} SVAMP \citep{patel2021svamp} is a benchmark of elementary math word problems built to defeat surface heuristics, with problem bodies that differ in structure from GSM8K's. We probe the answer-prefix residual on the subset the model solves correctly in direct mode ($n = 861$), using seed-disjoint cross-validation so that problems sharing a body never straddle the train/test split. Entity tokens are split into decorative and operand-attached by a mechanical attachment criterion rather than by hand. Two signatures are read on operand-resampled variants of the same problems ($n = 3{,}282$), which give the within-problem contrasts that GSM-Symbolic templates provide natively: operand-number decodability is measured against other instantiations of the same problem, and answer decodability against variants of the same problem with different answers, so that neither can be carried by narrative or topic.

\begin{figure}[tb]
  \centering
  \includegraphics[width=\columnwidth]{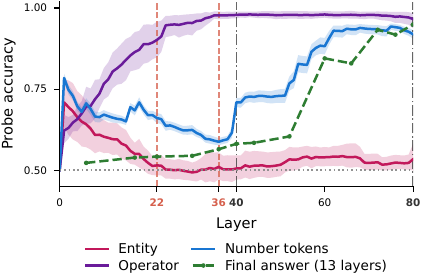}
  \caption{All four stage signatures reproduce on SVAMP at the layers found on
    GSM-Symbolic, with the same probe pipeline and no per-benchmark tuning
    (compare Fig.~\ref{fig:abstraction-main}b). Decorative-entity decodability
    falls from $0.71$ to $0.51$ at \Labs{}. Operator decodability is $0.90$ at
    \Labs{} and $0.98$ from \Lform{} onward. Operand numbers step to $0.71$ at \Lbind{}. The final answer rises
    from $0.58$ at \Lbind{} to $0.95$ at \Lout{}. The entity and operator curves
    are read on SVAMP itself. The operand-number and final-answer curves are
    read on the operand-resampled variants, so both are within-problem
    contrasts: negatives are other variants of the same problem. The
    final-answer probe comes from a separate diagnostic evaluated at $13$
    sampled layers, which is what the markers indicate.}
  \label{fig:svamp-transfer}
\end{figure}

Figure~\ref{fig:svamp-transfer} reports the comparison. Decorative-entity decodability peaks before \Labs{} and falls to chance at \Labs{}, operator decodability is high by \Labs{} and stays elevated while the plan forms and operands bind, operand numbers hold a plateau through L39 and step up at \Lbind{}, and answer decodability rises late and saturates at \Lout{}. The layer indices at which the four transitions occur are the ones identified on GSM-Symbolic, not re-estimated on SVAMP.

\paragraph{Beyond math: a proof of concept on PhantomWiki.} PhantomWiki \citep{gong2025phantomwiki} generates fictional biographical corpora together with multi-hop kinship questions over them. The task computes with different tokens than math does: the operations are relation words (\emph{mother}, \emph{husband}, and so on) and the values passed from one reasoning step to the next are person names. We generate universes and probe the answer-prefix residual on correctly answered questions ($n = 4{,}672$), with universe-disjoint cross-validation so that no universe appears in both train and test.

Three readouts are informative. \emph{Relation words} are decodable at $0.98$ by \Labs{} and remain at $0.96$ at \Lbind{}, the same profile operators show in math: the operation vocabulary is extracted early and retained while the rest of the pipeline runs. \emph{Chain membership}, whether a given name lies on the reasoning chain, is measured under a matched contrast in which positives are chain intermediates and negatives are uninvolved names drawn from the same article, matched on presence, adjacency, and difficulty, so that only chain membership differs. Under that contrast the probe sits at chance through L38 ($0.50$--$0.53$), rises sharply to $0.59$ at L39, and slowly reaches $0.66$ by \Lout{}. The \emph{answer entity} does not saturate to $1.00$ until \Lout{}.

The onset of both name signatures at L39 and their completion at \Lout{} sit at the boundaries of the math pipeline, even though the task is retrieval rather than arithmetic. The stage contents are task-shaped and the stage structure recurs: kinship questions have no decorative narrative to strip, and because each hop of a retrieval chain fetches and resolves at once, Operand Binding and Computation are not separated as cleanly as in math but compress into a single resolution cascade. We report this as a proof of concept rather than a full replication.

\end{document}